\documentclass[preprint, 12pt]{elsarticle}
\usepackage{graphicx}
\usepackage{epstopdf}
\usepackage{float}
\usepackage{caption}
\usepackage{booktabs}
\usepackage{multirow}
\usepackage{makecell}
\usepackage{xurl}
\usepackage{amssymb}
\usepackage{amsmath}
\usepackage{array}
\usepackage{bm}

\begin{document}

\begin{frontmatter}	

\title{Bid Optimization using Maximum Entropy Reinforcement Learning}

\author[uestc1]{Mengjuan Liu\corref{cor1}}
\ead{mjliu@uestc.edu.cn}

\author[uestc1]{Jinyu Liu}

\author[uestc1]{Zhengning Hu}

\author[uestc1]{Yuchen Ge}

\author[uestc1]{Xuyun Nie}

\cortext[cor1]{Corresponding author: Mengjuan Liu}
\small \address[uestc1]{Network and Data Security Key Laboratory of Sichuan Province, University of Electronic Science and Technology of China, Chengdu, 610054, China}

\begin{abstract}
Real-time bidding (RTB) has become a critical way of online advertising. In RTB, an advertiser can participate in bidding ad impressions to display its advertisements. The advertiser determines every impression’s bidding price according to its bidding strategy. Therefore, a good bidding strategy can help advertisers improve cost efficiency. This paper focuses on optimizing a single advertiser’s bidding strategy using reinforcement learning (RL) in RTB. Unfortunately, it is challenging to optimize the bidding strategy through RL at the granularity of impression due to the highly dynamic nature of the RTB environment. In this paper, we first utilize a widely accepted linear bidding function to compute every impression’s base price and optimize it by a mutable adjustment factor derived from the RTB auction environment, to avoid optimizing every impression’s bidding price directly. Specifically, we use the maximum entropy RL algorithm (Soft Actor-Critic) to optimize the adjustment factor generation policy at the impression-grained level. Finally, the empirical study on a public dataset demonstrates that the proposed bidding strategy has superior performance compared with the baselines.
\end{abstract}

\end{frontmatter}

\section{Introduction}
\label{Introduction}
Online advertising has developed into the most primary way of ad delivery in recent years \cite{ref1}. As a new marketing channel of online advertising, real-time bidding (RTB) has received extensive attention from industry and academia since it significantly improves the efficiency and transparency of the online advertising ecosystem. Figure 1 illustrates the typical process of an advertiser buying an ad impression through RTB \cite{ref2}. When a user browses a web page, the script of the ad slot embedded on the page will initiate a bidding request to the ad exchange (ADX). Then the ADX delivers the bidding request to the connected demand-side platforms (DSPs). The bidding agent running on each DSP calculates the bidding price of the auctioned impression for each advertiser based on its utility \cite{ref3}. The highest bidding price within each DSP is fed back to the ADX, and the ADX determines the final winner according to the generalized second pricing (GSP) mechanism \cite{ref4}. The winning notice is then sent to the winner, and its advertisement will be displayed to the user on the web page. Usually, the DSP will track the user’s click or conversion behavior to optimize the advertiser’s utility estimator and bidding strategy \cite{ref5}.\par
\begin{figure}[t]
\centering
\includegraphics[width=\textwidth]{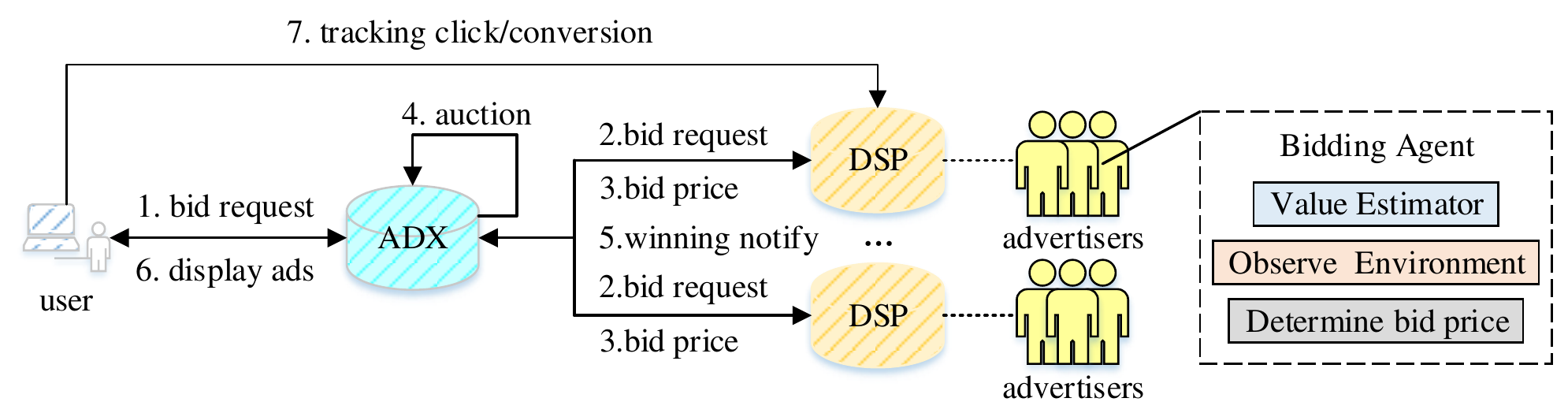}
\captionsetup{font={footnotesize}}
\caption{Typical Process of an Ad Delivery in RTB}
\label{fig:fig1}
\vspace{-0.10in}
\end{figure}
In RTB, each advertiser can dynamically adjust the bidding price based on the utility of the ad impression to itself. Therefore, an advertiser can maximize revenue under a limited budget by optimizing its bidding strategy. This paper focuses on optimizing a single advertiser’s bidding strategy. The advertiser’s revenue in RTB usually refers to the user’s click or conversion behavior after the ad is displayed \cite{ref6}. The more clicks (conversions) the advertiser gets in an ad delivery period, the greater the revenue. Therefore, the goal of the bidding strategy is to make a bidding decision for each auctioned ad impression so that a single advertiser can get the maximum number of clicks (or conversions) in an ad delivery period. To achieve this goal, the bidding agent should determine the bidding price based on the value of the impression to the advertiser, conforming to the optimal auction theory \cite{ref2}. Usually, the bidding agent in RTB uses the predicted click-through rate (pCTR) to measure the value of the impression and makes a bidding decision based on pCTR. So we can formalize the optimal bidding strategy of a single advertiser as the formula (1), maximizing the total value of purchased ad impressions in an ad delivery period under a given budget. \par
\begin{equation}
\begin{split}
b^*(i) & =\underset{b(\cdot)}{max}\sum_{i=1}^Tw(i)\cdot v(i) \\
s.t. &\sum_{t=1}^T w(i)\cdot cost(i)\leq B
\end{split}
\label{equ:equ1}
\end{equation}\par
In the formula (1), $v(i)$ is the estimated value of an ad impression \cite{ref7}, and $w(i)$ is a binary indicator whether the ad impression is purchased successfully using the bidding strategy $b(\cdot)$. The optimal bidding strategy $b^*(\cdot)$ maximizes the total value of all purchased ad impressions in the whole delivery period. In the constraint, $cost(i)$ refers to the cost of purchasing ad impression $i$, and the total cost of purchasing all impressions should not exceed the given budget $B$.\par
Among the existing bidding strategies, linear bidding strategy (called LIN) is one of the most representative schemes. It first designs a bidding function with pCTR as the variable and learns the parameters from the samples of historical periods via maximizing the total pCTR of all purchased impressions. Linear bidding strategy is static, simple, and easy to deploy, so it is widely employed on many DSP platforms. Unfortunately, the static bidding strategy does not work well when the RTB environment changes significantly between the historical and the new ad delivery periods. In the static bidding strategy, the bidding price is only related to the pCTR of an ad impression and cannot adapt to changes of the RTB auction environment. Intuitively, an ideal bidding strategy ought to be associated with the pCTR of ad impression and the real-time auction environment, such as the advertiser’s available budget, remaining life time, and the intensity of market competition.\par
To make bidding price change with the RTB environment, the authors in \cite{ref8} proposed using reinforcement learning (RL) \cite{ref9} to learn the optimal bidding strategy (called RLB). Through RL, the bidding agent considers not only the immediate reward from a single ad impression but also the cumulative benefit from all purchased impressions during the whole ad delivery period \cite{ref10}. However, RLB is a model-based RL bidding strategy that needs to establish the state transition probability matrix. When the number of states reaches billions, establishing this matrix requires substantial computational and storage costs, which fails to deploy in real RTB applications \cite{ref11}. A promising improvement is to use the latest model-free RL (DQN) algorithm to model bidding decisions \cite{ref12}. The new bidding strategy (called DRLB) not only enables bidding price to be correlated with both the impression’s pCTR and RTB environment, but also does not need to compute the state transition probability matrix. Yet, the experimental results demonstrate that DRLB does not outperform LIN and RLB. Its performance depends heavily on the initial value of the bidding factor and the design of discrete action space (the detailed explanation refers to Section 2).\par
Indeed, it is still challenging for the bidding agent to learn the optimal bidding strategy at the impression-grained level using the model-free RL algorithm. The difficulties are as follows. Firstly, the bidding agent located on the DSP only obtains incomplete information about the RTB environment. For example, it neither knows the lost ad impressions’ market prices nor how many advertisers participate in bidding the impression. The agent only represents the state by some observable statistical metrics, which may cause the bidding agent to make a non-optimal bidding decision. Secondly, RL guides the bidding agent to learn the optimal bidding strategy through the reward mechanism. But defining an appropriate immediate reward is a tough job in RTB. For instance, RLB directly uses the pCTR of the purchased ad impression as the immediate reward, which easily guides the agent to learn to bid with an exorbitant price to obtain the pCTR without considering the cost. As a result, the agent buys plenty of low value impressions, resulting in budget waste.\par
Thirdly, the GSP auction is widely adopted in RTB, which means the advertiser who bids greater or equal to the second-highest bidding price wins the ad impression but is only charged the second-highest price. That is to say, for a single ad impression, multiple bidding prices are corresponding to the same benefit and cost, expressing there are multiple optimal actions at every state. However, the current mainstream model-free RL algorithms (such as DQN \cite{ref12} and DDPG \cite{ref13}) are deterministic, in which there is only one optimal action at every state, and they only maximize the probability of the optimal action being selected during the learning process. This inconsistency may prevent the bidding agent from learning the optimal bidding price at each state.\par
To overcome the above difficulties, we design a new bidding function to calculate the bidding price for a single ad impression in this paper. As shown in formula (2), it contains a base price and an adjustment value. The base price is computed by a linear bidding function derived by a heuristic algorithm. The adjustment value is obtained by multiplying the optimal adjustment factor (generated by the RL agent) by the range of bidding adjustment. Thus, through formula (2), the bidding agent no longer selects an optimal bidding price from a discrete price space but generates a continuous adjustment factor with a given range. This bidding function ensures that the bidding price does not deviate too much from its estimated value and can be adjusted according to the real-time RTB environment.\par
\begin{equation}
\begin{split}
b(i) & = b_{LIN}(i)+\Delta b(i) \\
& =\underbrace{b_{LIN}(i)}_{base\ price}+\underbrace{a_i}_{adjustment\ factor}\times \underbrace{\min({price}_{\max}-b_{LIN}(i), b_{LIN}(i)-{price}_{\min})}_{bidding\ adjustment\ range}
\end{split}
\label{equ:equ2}
\end{equation}\par
Furthermore, we model the adjustment factor decisions of ad impressions in an ad delivery period as an MDP \cite{ref14}. Therefore, the RL agent’s task is to learn the optimal adjustment factor generation policy. Specifically, we introduce a maximum entropy RL algorithm, Soft Actor-Critic (SAC) \cite{ref15}, to generate each ad impression’s adjustment factor. Unlike the deterministic RL algorithms, SAC has a strong self-learning ability and can balance the probabilities of multiple optimal actions being selected in the same state. Thus, our bidding strategy can solve the third difficulty mentioned above effectively using SAC. Besides, SAC expands the scope of the agent to explore the optimal action to avoid falling into the local optimum. The contributions of our work can be summarized as follows:\par
\begin{itemize}
\item We design a new bidding function by improving the linear one, which considers both the impression’s estimated value and the real-time RTB environment. Specifically, we introduce an adjustment factor into the bidding function, which can be adjusted according to the real-time RTB environment dynamically. Thus, the RL agent in our strategy only needs to learn the optimal adjustment factor generation policy, avoiding generating the bidding price directly. This greatly reduces the difficulty of the RL agent learning the optimal action generation policy.\par
\item 	To generate the optimal adjustment factor for each impression, we model the adjustment factor decisions as an MDP and optimize the adjustment factor generation policy through SAC. Using SAC not only overcomes the problem of multiple optimal actions for each impression but also expands the exploration space of the optimal action to avoid falling into the local optimum. Concretely, we redefine the state representation and reward function in the MDP for enabling the RL agent to understand the optimization objective better. It is the first bidding strategy at an impression-grained level through stochastic reinforcement learning to the best of our knowledge.\par
\item 	We evaluate the proposed scheme and several baselines on a benchmark dataset, and the results demonstrate our method outperforms other baselines. Specially, we are the first to quantitatively validate the impact of the dynamic RTB environment on the performance of static bidding strategy (e.g., LIN). Furthermore, we discuss the detailed differences between our approach and LIN.\par
\end{itemize}
\section{Related Work}
\label{Related Work}
Recently, research on bidding strategies in RTB mainly focuses on static strategy, designing a linear or non-linear bidding function about impression’s pCTR, derived by heuristics algorithm or optimization method from the historical data. Then, the bidding agent directly uses this learned bidding function to bid for each ad impression in the new ad delivery period. The representative static bidding strategies are LIN \cite{ref16} with linear function and ORTB \cite{ref17} with non-linear function, as shown in formula (3) and formula (4). Here $base\_bid$ is a fixed base bid, and $avg\_pctr$ is the average value of all ad impressions' pCTRs on the training set. In LIN, we set the base bid from 1 to 300 (increased one by each time) and calculate each ad impression's bidding price according to (3). Finally, the base bid that maximizes the total clicks on the training set is recorded as the optimal base bid and is used in the new ad delivery period. In ORTB, the parameters $c$ and $\lambda$ are both learned from the training set. So, during a new ad delivery period, the bidding price in ORTB only depends on the pCTR of the impression.\par
\begin{align}
bid_{LIN}(i)&=pctr(i)\times \frac{base\_bid}{avg\_pctr} \\
bid_{ORTB}(i)&= \sqrt{\frac{c}{\lambda} pctr(i)+c^2} - c
\label{equ:equ3, equ:equ4}
\end{align}\par
The static bidding strategies described above have obvious flaws, as bidding price is only related to the pCTR of the ad impression in the new ad delivery period, and it cannot adapt to changes in the RTB environment. For example, when the market competition intensifies, the bidding agent should appropriately increase the bidding price and vice versa. Intuitively, we hope that strategy can dynamically allocate the budget to all ad impressions during the entire delivery period to maximize the total clicks or the cumulative pCTR of the purchased ad impressions. This requires the bidding agent to adjust the bidding strategy dynamically according to the real-time RTB environment. Reinforcement learning may be a promising solution to accomplish this task because it can achieve excellent decision-making. In RTB, the RL agent considers not only the benefit of the single ad impression but its impact on long-term profits.\par
RLB is a typical bidding strategy based on a model-based RL framework \cite{ref8}. The author creatively modeled the bidding decision for each ad impression within an ad delivery period as a sequential MDP \cite{ref18}. The entire RTB system is regarded as the environment. Each impression reaches the DSP triggers state transfer. The bidding agent first observes the state from the RTB environment and then selects an action from the pre-designed discrete action (price) space [0, 1, ..., 300] ($10^{-3}$ Chinese FEN) as its bidding price. If the agent wins the ad impression, the environment will feedback the pCTR of the winning ad impression as the immediate reward. RLB utilizes a dynamic programming algorithm to optimize the optimal action selection policy based on a model-based RL model. The disadvantage of model-based RL is that it is necessary to establish a state transition probability matrix. For a real-world RTB system with billions of ad impressions, the establishment of this matrix requires huge computational and storage overhead, which makes it impossible to deploy on a real DSP.\par
As an improvement, researchers seek to solve the MDP by using model-free RL algorithms. Model-free RL is particularly suitable for scenarios where the agent cannot obtain complete environmental information. It does not need to establish a state transition probability matrix. Instead, the agent in model-free RL learns the optimal action selection policy from experience. However, not as expected, it is challenging to use typical model-free RL algorithms such as DQN and DDPG to learn the optimal action selection policy for a single ad impression. In extreme cases, the bidding agent can only learn to bid with a single high/low price or fall into a locally optimal solution for a long time and cannot explore a larger price space. We have analyzed the causes of this situation in detail in the introduction.\par
Because of the above problems, the authors in \cite{ref11} proposed DRLB, which gave up bidding directly on a single ad impression but introduced a time-related adjustment factor based on the linear bidding function. As shown in formula (5), where $bid(i,t)$ represents the bidding price for \textit{i}-th ad impression in \textit{t}-th time slot, $pctr(i)$ represents the estimated value of the ad impression $i$, $\lambda(t)$ represents the scale bidding factor in each time slot. The greedy algorithm is used to obtain the initial bidding factor $\lambda(0)$. By observing each time slot's RTB environment, the bidding agent selects an optimal action as the adjustment factor for the new time slot and adjusts the bidding price of the ad impression that reaches in this time slot. In DRLB, the bidding price is related to the pCTR of the impression and the RTB environment of the current time slot. DRLB uses the value-based model-free RL algorithm (DQN) to learn the bidding factor's adjustment value for each time slot.\par
\begin{equation}
\begin{split}
bid(i, t) &= pctr(i)/\lambda(t) \\
\lambda(t) = \lambda(t&-1)\times (1+\beta_\alpha(t)) \\
\beta_\alpha(t) \in \{-8\%, -3\%, -1\%, 0&\%, 1\%, 3\%, 8\%\}\ t=1,2,\cdots,T
\end{split}
\label{equ:equ5}
\end{equation}\par
The problem with DRLB is how to design a suitable discrete action space and the initial value of the bidding factor (the value of the first time slot). Therefore, based on DRLB, literature \cite{ref19} generates the optimal adjustment factor for each time slot by introducing a deterministic policy algorithm — Twin Delayed Deep Deterministic policy gradient (TD3). Unlike the discrete action space used by DRLB, the action space designed by \cite{ref19} is a continuous value in (-1, +1).\par
In this paper, we follow the idea of DRLB, which does not directly model the bidding decision, but models the bidding adjustment factor decision. Further, we implement the adjustment of a single ad impression. The bidding agent learns the bidding adjustment factor of the current state and adjusts the bidding price derived from a static bidding strategy. Simultaneously, for the agent to better understand the phenomenon of multiple optimal actions led by the GSP mechanism, we use the RL algorithm based on maximum entropy (SAC) to train the bidding agent to make the optimal adjustment factor generation at each ad impression. Besides, SAC also overcomes the problem of narrow exploration scope caused by deterministic policy. As far as we know, our scheme is the first to apply the maximum entropy stochastic policy RL algorithm to optimize RTB bidding strategy. Table 1 summarizes the differences between several representative bidding strategies and ours.\par
\begin{table}[t]
\Huge
\renewcommand\arraystretch{1.5}
\centering
\caption{Characteristics of Typical Bidding Strategies}
\resizebox{\textwidth}{!}{%
\begin{tabular}{ccccccc}
\toprule
\toprule
\multicolumn{1}{c}{\textbf{Strategy}} & \multicolumn{1}{c}{\textbf{Static/Dynamic}} & \multicolumn{1}{c}{\textbf{Method}} & \multicolumn{1}{c}{\textbf{Adaptiveness}} & \multicolumn{1}{c}{\textbf{Granularity}} & \multicolumn{1}{c}{\textbf{Action}} & \multicolumn{1}{c}{\textbf{Action Space}} \\ 
\midrule
\multicolumn{1}{c}{\textbf{LIN}} & Static & Linear/Heuristic & Not support & Impression & / & / \\
\multicolumn{1}{c}{\textbf{ORTB}} & Static & \makecell*[c]{Non-linear\\Optimization} & Not support & Impression & / & / \\
\multicolumn{1}{c}{\textbf{RLB}} & Dynamic & \makecell*[c]{Model-based RL\\Dynamic programming} & Support & Impression & Bidding price & Discrete value \\
\multicolumn{1}{c}{\textbf{DRLB}} & Dynamic & Model-free RL/DQN & Support & Time slot & \makecell*[c]{Bidding factor's \\ regulating value} & Discrete value \\
\multicolumn{1}{c}{\textbf{OURS}} & Dynamic & Model-free RL/SAC & Support & Impression & Bidding factor & Continuous value \\
\bottomrule
\bottomrule
\end{tabular}}%
\label{tab:tab1}
\vspace{-0.10in}
\end{table}
\section{Problem and Formulation}
\label{Problem and Formulation}
In this paper, to avoid directly learning the optimal bidding price for individual impression, we first design a new bidding function, as shown in formula (2). Its first part is the base price, and the second part is the optimal adjustment value generated by the RL agent according to the real-time RTB auction environment. To be specific, we use LIN \cite{ref16} to generate the base price and obtain the optimal adjustment factor $a_i$ for each ad impression by using SAC-based RL algorithm. It is noted that we use $\min (price_{\max}-b_{LIN}(i),b_{LIN}(i)-price_{\min})$ to limit the range of the adjustment value, where $price_{\max}$ and $price_{\min}$ are the maximum and minimum bidding prices for an impression preset by the advertiser.\par
We regard an ad delivery period as an episodic process in this paper. The decision-making of bidding adjustment factors for all sequential ad impressions in the entire delivery period is modeled as a MDP. In the model-free RL framework, the agent observes a state directly from the environment, so it is no longer necessary to calculate the state transition probability matrix. MDP can be represented by $\langle \mathcal{S}, \mathcal{A}, \mathcal{R}\rangle$, where $\mathcal{S}$ is the state space $(s_t \in \mathcal{S})$, and $\mathcal{A}$ is the action space $(a_t \in \mathcal{A})$. And $\mathcal{R}$ is the reward function that decides the immediate reward received after taking action $a_t$ under the state $s_t$. The interaction process between the bidding agent and the RTB environment can be shown in Figure 2.\par
\begin{figure}[t]
\centering
\includegraphics[width=0.6\textwidth]{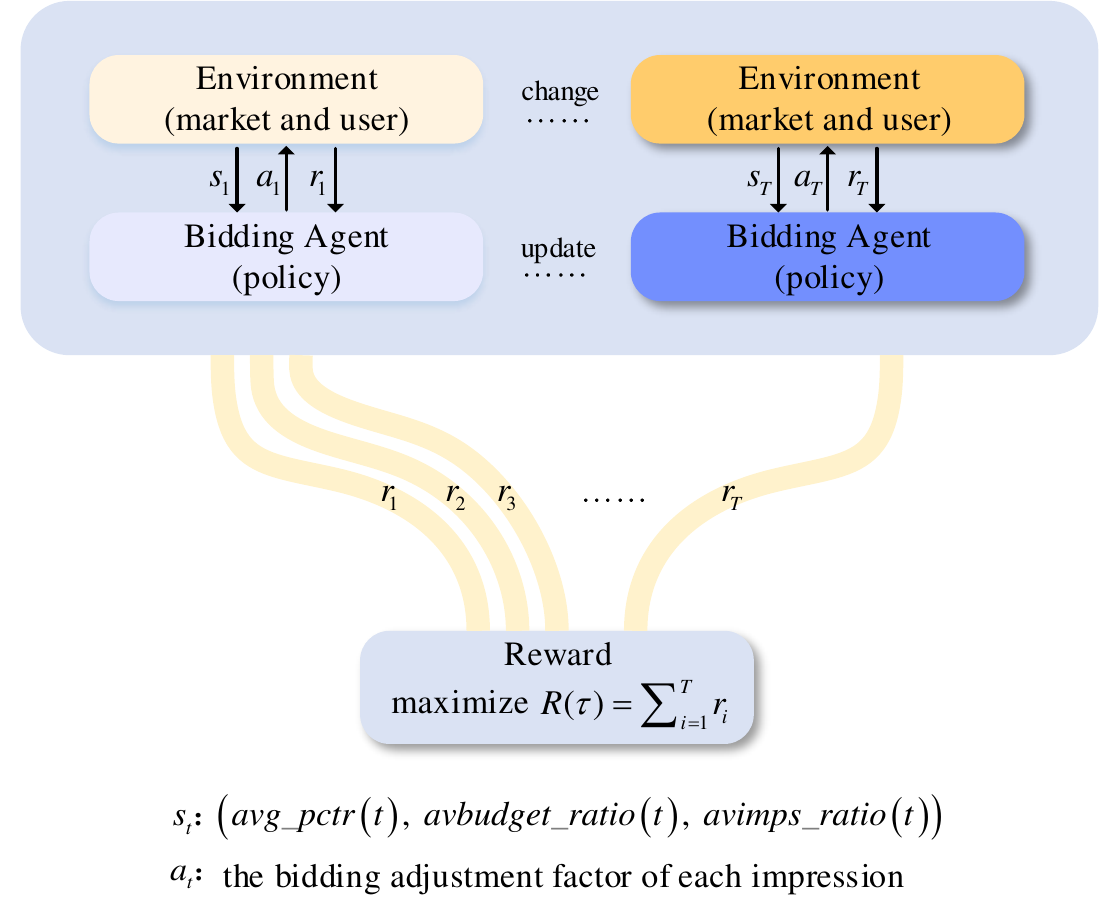}
\captionsetup{font={footnotesize}}
\caption{Interaction process between the bidding agent and the RTB environment}
\label{fig:fig2}
\vspace{-0.10in}
\end{figure}
Next, we describe the key components of our MDP as follows:\par
\noindent \textbf{State}: The bidding agent observes the state from the environment and uses statistical information to represent the state, as shown in (6): \par
\begin{equation}
s_t=(avg\_pctr(t), avbudget\_ratio(t), avimps\_ratio(t))
\label{equ:equ6}
\end{equation}
where each parameter is described as:\par
\begin{itemize}
\item $avg\_pctr(t)$: the average pCTR of the received impressions, as defined in formula (7). It reflects the average quality of ad impressions in the new ad delivery period.\par
\begin{equation}
avg\_pctr(t)=\frac{1}{t}\sum_{i=1}^tpctr(i)
\label{equ:equ7}
\end{equation}
\item $avbudget\_ratio(t)$: the ratio of the advertiser’s available budget to the allocated budget, as defined in formula (8). In our algorithm, the impressions of the entire delivery period are divided into several slots for every 1000 impressions. When a new slot begins, the agent allocates the budget for the incoming slot with CPM $\times$ 1000, where CPM is the average market price of all purchased impressions in the new delivery period. The cost of the current slot should not exceed its budget.\par
\begin{equation}
avbudget\_ratio(t)=avbudget(t)/budget(t)
\label{equ:equ8}
\end{equation}
\item $avimps\_ratio(t)$: the ratio of the number of remaining ad impressions to 1000 in the slot at which the \textit{t}-th ad impression arrives, as shown in formula (9).\par
\begin{equation}
avimps\_ratio(t)=avimps\_num/1000
\label{equ:equ9}
\end{equation}
\end{itemize}
\noindent \textbf{Action}: We defined a new bidding function to calculate the bidding price for each ad impression. Based on this design, the agent's task is to generate action (adjustment factor) $a_t$ suitable for the current state at the impression-grained level. The agent first generates a probability distribution about action according to the state and then samples an action randomly on this distribution. Finally, the action value is restricted to (-1, 1) using the tanh function \cite{ref20}.\par
\noindent \textbf{Reward}: The reward function is defined as (10). Different from the definition of RLB and DRLB, we combine our bidding, linear bidding (reflecting the value of ad impression), and the actual market price together to determine the immediate reward. As shown in formula (10), $b_{LIN}(t)$ represents the linear bidding of the \textit{t}-th ad impression, $market(t)$ represents the market price (second-highest price) of this ad impression, and $b(t)$ represents the bidding price according to our bidding.\par
\begin{scriptsize}
\begin{equation}
\displaystyle
reward(t)=
\begin{cases}
pctr(t),&if\ b_{LIN}(t)<market(t)\ \& \ b(t)\geq market(t) \vspace{1ex}\\ 
pctr(t)\times \frac{avbudget(t)}{\left| b(t)-b_{LIN}(t)\right|+1},&if\ b_{LIN}(t)\geq market(t)\ \& \ b(t)\geq market(t) \vspace{1ex}\\
pctr(t)\times (a_t-1),&if\ b_{LIN}(t)<market(t)\ \& \ b(t)<market(t) \vspace{1ex}\\
pctr(t)\times a_t,&if\ b_{LIN}(t)\geq market(t)\ \& \ b(t)<market(t) \vspace{1ex}\\
-pctr(t),&if\ avbudget(t)<b(t)\\
\end{cases}
\label{equ:equ10}
\end{equation}
\end{scriptsize}
\noindent The specific rules are designed as:\par
\begin{itemize}
\item If our bidding can win the ad impression, and linear bidding cannot win, the environment feedbacks a positive reward.\par
\item If both our bidding and linear bidding can win the ad impression, the environment also feedbacks a positive reward. Because the market price of the impression is roughly positively correlated with its value, this design makes our bidding close to the linear bidding based on the impression’s value.\par
\item If neither our bidding nor the linear bidding can win the ad impression, the environment feedbacks a negative reward. This design is to enable $a_t \to 1$ to increase our bidding price to win this ad impression as much as possible.\par
\item If the linear bidding can win the auction, and our bidding cannot win, the environment feedbacks a negative reward. This design is to enable $a_t \to 1$ to increase our bidding price to win this ad impression.\par
\item If the budget of the current slot is insufficient or it is spent out in advance, considering that all arriving subsequent ad impressions cannot be purchased. In this case, the environment feedbacks a negative reward. Because the budget spent out in advance will cause the advertiser to lose all subsequent impressions at the slot, we use the negative rewards to make the bidding agent avoid such situation as much as possible. \par
\end{itemize}

\section{Solution based on SAC}
\label{Solution based on SAC}
Different from the previous RL bidding strategy, we use the stochastic policy algorithm (SAC) to optimize adjustment factor generation policy. SAC is based on the Actor-Critic framework, drawing on the structural design in DDPG \cite{ref13}, setting up a Target network and an Eval network to enhance the stability of the model, and also using two Q networks to solve the problem of bias caused by overestimation in RL following the technology in TD3 \cite{ref21}. Additionally, SAC introduces the policy entropy proposed in Soft Q learning \cite{ref22} to balance the stochasticity of action selection to deal with the problem of multiple optimal actions for each ad impression caused by the GSP mechanism in RTB. \par
In the basic RL framework, we use the Temporal-Difference (TD) method to optimize the strategy, and the optimization goal is represented by the formula (11), where $r(s_t,a_t)$ represents the immediate reward brought by performing the action $a_t$ under the state $s_t$.\par
\begin{table}[t]
\normalsize
\centering
\caption{Key Parameters of Figure 3}
\resizebox{\textwidth}{!}{%
\begin{tabular}{cc}
\toprule
\toprule
\multicolumn{1}{c}{\textbf{Variable}} & \multicolumn{1}{c}{\textbf{Description}} \\ 
\midrule
$\alpha$ & \makecell*[l]{The temperature parameter determines the relative importance\\ of the entropy term versus the reward, thus controlling the\\ stochasticity of the optimal policy. }\\
$\hat{a}_{t+1}\sim \pi_\phi(s_{t+1})$ & \makecell*[l]{Generate action distribution on the state $s_{t+1}$ through the Actor\\ network and randomly sample to generate action $\hat{a}_{t+1}$.} \\
$\log \pi_\phi(a|s)$ & \makecell*[l]{The entropy of action $a$ generated by the Policy network on the\\ state $s$, where $\phi$ represents the parameters of the Policy network.} \\
$Q_{\theta_i}(s_t,a_t),i=1,2$ & \makecell*[l]{An Eval Critic network used to calculate the Q value upon the \\state $s_t$ and the action $a_t$, where $\theta_i$ means the parameters of the\\ Eval Critic network $i$.}\\
\bottomrule
\bottomrule
\end{tabular}}%
\label{tab:tab2}
\vspace{-0.10in}
\end{table}
\begin{equation}
\pi^*=\mathop{\arg \max}_\pi \mathbb{E}_{\left(s_t,a_t\right)\sim \rho_\pi}[\sum\nolimits_{t=1}^Tr\left(s_t,a_t\right)]
\label{equ:equ11}
\end{equation}
SAC introduces the policy entropy, which is defined as $\mathcal{H}(P)=\underset{x\sim P}{E}[-\log P(x)]$. $P(x)$ represents the probability distribution of $x$. Entropy term usually represents the degree of confusion in action selection. Here we hope to enhance the self-learning ability of this model. That is, let actions with the same reward have the same probability of being selected as much as possible. Therefore, the optimization goal is modified to formula (12), where $\pi(\cdot|s_t)$ represents the probability distribution of actions based on the RL policy under the current environment state $s_t$, $\alpha$ is the temperature parameter used to balance entropy and reward, thereby controlling the stochasticity of the optimal strategy.\par
\begin{equation}
\pi^*=\mathop{\arg \max}_\pi \mathbb{E}_{\left(s_t,a_t\right)\sim \rho_\pi}[\sum\nolimits_{t=1}^Tr\left(s_t,a_t\right)+\alpha\mathcal{H}(\pi(\cdot|s_t))]
\label{equ:equ12}
\end{equation}
According to \cite{ref23}, we automatically adjust the temperature parameter when it is not less than the minimum policy entropy threshold and maximize the original reward. The optimization goal is defined in the formula (13), where $\mathcal{H}_0$ is the preset minimum policy entropy threshold, we also set $\mathcal{H}_0=-\dim(\mathcal{A})$ according to \cite{ref23}. The entire process is shown in Figure 3, and we describe some critical parameters in detail in Table 2.\par
\begin{equation}
\underset{\pi_0,\cdots,\pi_T}{\max}\mathbb{E}[\sum\nolimits_{t=0}^Tr(s_t,a_t)] \ s.t.\forall t,\mathcal{H}(\pi_t)\geq \mathcal{H}_0
\label{equ:equ13}
\end{equation}\par
\begin{figure}[!]
\centering
\includegraphics[width=\textwidth]{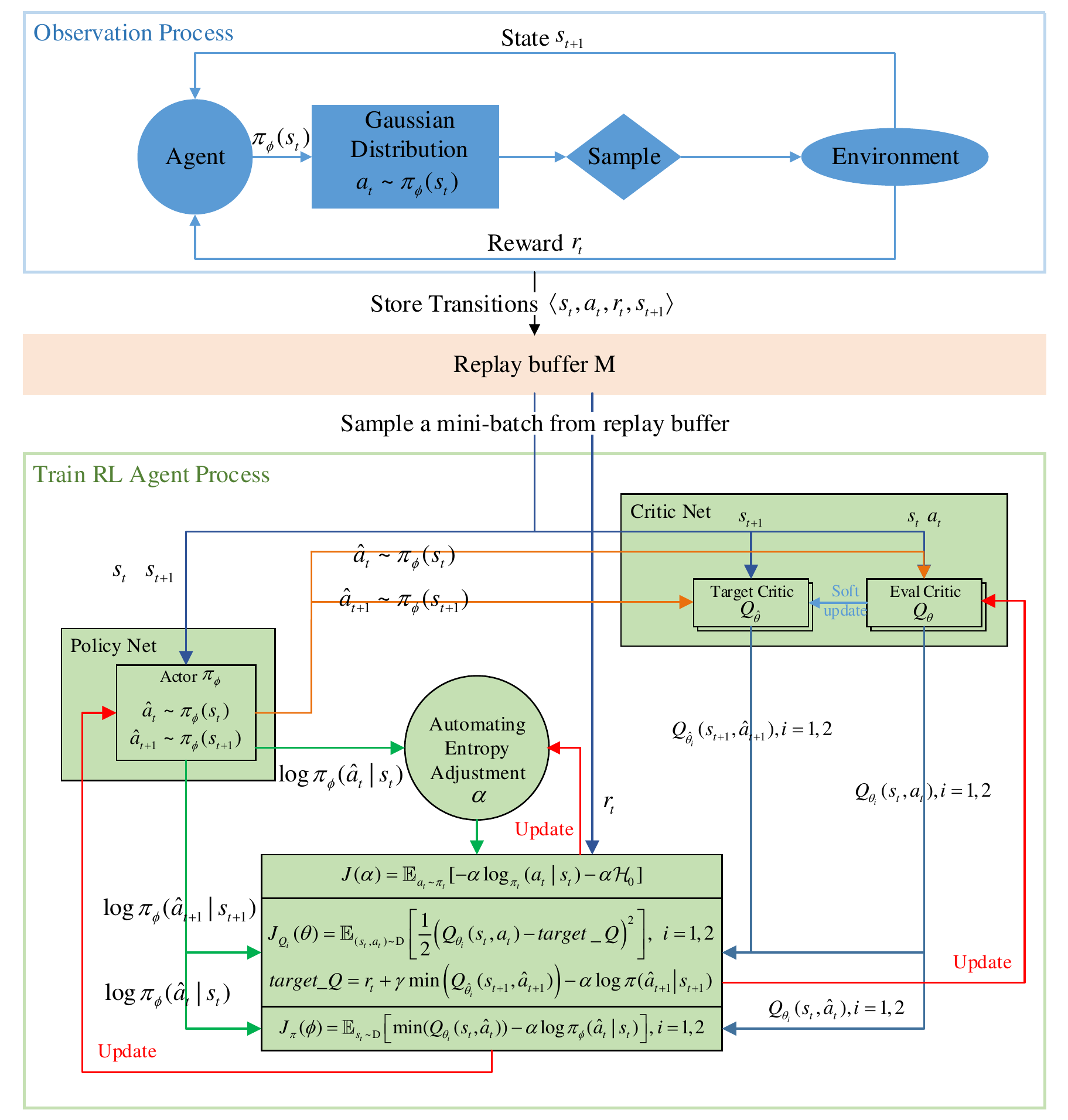}
\captionsetup{font={footnotesize}}
\caption{Structure of our learning algorithm based on SAC}
\label{fig:fig3}
\vspace{-0.10in}
\end{figure}
As shown in Figure 3, there are \textbf{observation} and \textbf{training} processes in our algorithm. Before training, we first need to initialize a replay buffer, recorded as $M$. The replay buffer is a fixed-sized cache. Transitions were sampled from the interaction process between agent and environment. Specifically, upon each ad impression's arrival, the agent generates a Gaussian distribution $\pi_\phi(s_t)$ according to the current state $s_t$. And then samples a bidding adjustment factor $a_t\sim\pi_{\phi}(s_t)$ from the distribution to adjust the base price of each ad impression according to formula (2). After that, the agent uses our bidding and linear bidding to compute the immediate reward based on the formula (10). At the same time, the agent computes the environment’s statistics as the next state $s_{t+1}$. Finally, each transition $\langle s_t, a_t, r_t, s_{t+1}\rangle$ is stored in the replay buffer. Whenever $k$ new transitions are added to the replay buffer, the agent performs a training.\par
Then, enter the training process. As shown in Figure 3, the architecture consists of two parts: Policy Net and Critic Net. The Policy Net is the Actor network for generating each impression’s bidding adjustment factor. The Critic Net includes two Eval Critic networks and two Target Critic networks. The two Eval Critic networks are used to compute the Q value given a state and an action. And the two Target Critic networks help update two Eval Critic networks. In our algorithm, the parameters of Actor Net and Eval Critic networks are updated based on training; the parameters in Target Critic networks are updated from the corresponding networks in Eval Critic networks by using soft update.\par
Now, we introduce the details of the training. We use neural network to fit Actor and Critic Net. Whenever $k$ pieces of new transition are stored in the replay buffer, the agent will train for $L$ rounds. As shown in Algorithm 1, for each round, the agent first randomly sample $N$ transitions from the replay buffer to form a mini-batch for training. For each transition $\langle s_t, a_t, r_t, s_{t+1}\rangle$, $s_t$ is input into the Policy network to generate the probability distribution of the action, $\pi_\phi(s_t)$, where $\phi$ is the parameters of the Policy network. Then the agent samples an action $\hat{a}_t$ from the distribution $\pi_\phi(s_t)$ and feed it into two Eval Critic networks respectively to compute two Q values, recorded as $Q_{\theta_i}(s_{t}, \hat{a}_{t})$, where $i=1,2$. Here, $\theta_i$ are two Eval Critic networks’ parameters. $\alpha\log\pi_\phi(\hat{a}_t|s_t)$ is the entropy term, and the temperature parameter is automatically adjusted by formula (16) to control the stochasticity of the optimal strategy. So we can update the Policy network’s parameters by using unbiased gradient estimator proposed in SAC, as shown in formula (14). It is worth noting that the agent chooses the smaller Q value to update the Policy network to avoid overestimation.
\begin{equation}
J_\pi(\phi)=\mathbb{E}_{s_t\sim \rm{D}}\left[\min(Q_{\theta_i}(s_t,\hat{a}_t))-\alpha\log\pi_\phi(\hat{a}_t|s_t)\right],i=1,2
\label{equ:equ14}
\end{equation}\par
At the same time, the agent inputs $(s_t,a_t)$ into two Eval Critic networks respectively to compute two Q values, recorded as $Q_{\theta_i}(s_{t}, a_{t})$, where $i=1,2$. We update two Eval Critic networks’ parameters by minimizing TD error, as shown in formula (15). Where $r_t$ comes from the input $\langle s_t, a_t, r_t, s_{t+1}\rangle$, $\gamma$ is the discount factor, $Q_{\hat{\theta}_1}(s_{t+1}, \hat{a}_{t+1})$ and $Q_{\hat{\theta}_2}(s_{t+1}, \hat{a}_{t+1})$ are Q values computed by two Target Critic networks upon the next state $s_{t+1}$ and the action $\hat{a}_{t+1}$. Here, $\hat{a}_{t+1}$ is the action sampled from distribution $\pi_\phi(s_{t+1})$. As with the update of Actor network, in order to prevent overestimation, we use the smaller Q value to update the gradient. The entropy term is also considered in the update formula.\par
\begin{table}[!t]
\normalsize
\centering
\resizebox{\textwidth}{!}{%
\begin{tabular}{lr}
\toprule

\multicolumn{2}{l}{\textbf{Algorithm 1 }Learning algorithm in our bidding strategy}\\ 
\midrule
\multicolumn{2}{l}{Initialize $Q_{\theta_i},\ \pi_\phi$ with random parameters $\theta_i,i=1,2$ and $\phi$}\\
\multicolumn{2}{l}{Initialize target networks $\hat{\theta}_i\gets \theta_i$ and $\hat{\phi}\gets \phi$}\\
\multicolumn{2}{l}{Initialize empty replay buffer $M$}\\
\multicolumn{2}{l}{\textbf{for} $episode=1$ to $E$ \textbf{do}}\\
\multicolumn{2}{l}{\hspace{1em}\textbf{for} $t=1$ to $T$ \textbf{do}}\\
\multicolumn{2}{l}{\hspace{2em}Get the base price according to (3)}\\
\multicolumn{2}{l}{\hspace{2em}Observe state $s_t$ and get $a_t$ from $a_t\sim \pi_\phi(s_t)$}\\
\multicolumn{2}{l}{\hspace{2em}Execute $a_t$ to adjust the bidding price shown in (2)}\\
\multicolumn{2}{l}{\hspace{2em}Obtain the reward $r_t$ from (10) and observe next state $s_{t+1}$}\\
\multicolumn{2}{l}{\hspace{2em}Store transition $\langle s_t,a_t,r_t,s_{t+1}\rangle$ in $M$}\\
\multicolumn{2}{l}{\hspace{2em}\textbf{if} $t \mod k$ \textbf{do}}\\
\multicolumn{2}{l}{\hspace{3em}\textbf{for} $l=1$ to $L$ \textbf{do}}\\
\multicolumn{2}{l}{\hspace{4em}Sample a mini-batch of $N$ transitions $\langle s_t,a_t,r_t,s_{t+1}\rangle$ from $M$}\\
\hspace{4em}$\theta_i\gets \theta_i-\lambda_Q\hat{\nabla}_{\theta_i}J_Q(\theta_i),\ i = 1,2$ & $\triangleright$Update the Q-function parameters\\
\hspace{4em}$\phi\gets\phi-\lambda_\pi \hat{\nabla}_\phi J_\pi(\phi)$ & $\triangleright$Update the policy weights\\
\hspace{4em}$\alpha\gets\alpha-\lambda\hat{\nabla}_\alpha J(\alpha)$ & $\triangleright$Adjust temperature\\
\multicolumn{2}{l}{\hspace{4em}\textbf{if} $l \mod d$ \textbf{do}}\\
\hspace{5em}$\hat{\theta}_i\gets \tau\theta_i+(1-\tau)\hat{\theta}_i,\ i = 1,2$ & $\triangleright$Update target network\\
\multicolumn{2}{l}{\hspace{4em}\textbf{end if}}\\
\multicolumn{2}{l}{\hspace{3em}\textbf{end for}}\\
\multicolumn{2}{l}{\hspace{2em}\textbf{end if}}\\
\multicolumn{2}{l}{\hspace{1em}\textbf{end for}}\\
\multicolumn{2}{l}{\textbf{end for}}\\
\bottomrule
\end{tabular}}%
\label{tab:tab999}
\vspace{-0.10in}
\end{table}
\begin{equation}
\begin{split}
J_{Q_i}(\theta)&=\mathbb{E}_{(s_t,a_t)\sim \rm{D}}\left[\frac{1}{2}\left(Q_{\theta_i}(s_t,a_t)-target\_Q\right)^2\right],\ i=1,2\\
target\_Q&=r_t+\gamma\min\left(Q_{\hat{\theta}_i}(s_{t+1},\hat{a}_{t+1})\right)-\alpha\log\pi(\hat{a}_{t+1}|s_{t+1})
\end{split}
\label{equ:equ15}
\end{equation}\par
As with \cite{ref23}, we use the following formula (16) to calculate the gradient for $\alpha$.\par
\begin{equation}
J(\alpha)=\mathbb{E}_{a_t\sim\pi_t}\left[-\alpha\log_{\pi_t}(a_t|\pi_t)-\alpha\mathcal{H}_0\right]
\label{equ:equ16}
\end{equation}\par
As with \cite{ref13}, we use soft-update to update two Target Critic networks, as shown in formula (17), where $\tau$ is the update weight.\par
\begin{equation}
\hat{\theta}_i\gets \tau\theta_i+(1-\tau)\hat{\theta}_i,\ \ i = {1,2}
\label{equ:equ17}
\end{equation}\par
\section{Experimental Setup}
\label{Experimental Setup}
\subsection{Dataset}
We perform the experiments on a benchmark dataset — iPinYou, which comes from a well-known DSP company and contains logs of impressions, bids, clicks, and final conversions. In iPinYou, DSP adopts a fixed bidding strategy and bids 300 ($10^{-3}$ Chinese FEN) for each arriving ad impression. Then ADX determines the winner according to the GSP auction mechanism. The winner needs to pay the second-highest price to ADX to purchase the impression, which we call the second-highest price the market price. As a result, DSP cannot know the lost impression’s market price and click (conversion) behavior. So we only use the logs of winning impressions to construct the dataset in our experiments, which makes the number of ad impressions in the dataset far less than the actual number. In detail, we select three datasets with advertiser IDs of 1458, 3358, and 3427. Each dataset contains ten days of advertising logs (from 2013/6/6 to 2013/6/15). We use the data of the first seven days (6/6-6/12) as the training set and the data of the last three days (6/13-6/15) as the testing set. The statistics of the three datasets are shown in Table 3. We regard every day as an ad delivery period and set the daily budget separately. Therefore, the training set includes seven ad delivery periods, and the testing set includes three ad delivery periods.\par
\begin{table}[!t]
\centering
\caption{Statistics on the Training and Testing Sets}
\resizebox{\textwidth}{!}{%
\begin{tabular}{rcccccc}
\toprule
\toprule
\multicolumn{1}{c}{\multirow{2}[4]{*}{}} & \multicolumn{2}{c}{\textbf{1458}} & \multicolumn{2}{c}{\textbf{3358}} & \multicolumn{2}{c}{\textbf{3427}} \\
\cmidrule{2-7}
\multicolumn{1}{c}{} & \multicolumn{1}{c}{\textbf{Training Set}} & \multicolumn{1}{c}{\textbf{Testing Set}} & \multicolumn{1}{c}{\textbf{Training Set}} & \multicolumn{1}{c}{\textbf{Testing Set}} & \multicolumn{1}{c}{\textbf{Training Set}} & \multicolumn{1}{c}{\textbf{Testing Set}} \\
\midrule
\textbf{Imps} & 3083056 & 614638 & 1742104 & 300928 & 2593765 & 536795 \\
\textbf{Clicks} & 2454  & 515   & 1358  & 260   & 1926  & 366 \\
\textbf{Cost} & 212400241 & 45216454 & 160943087 & 34159766 & 210239915 & 46356518 \\
\textbf{CTR(10$^{-3}$)} & 0.7959 & 0.8378 & 0.7795 & 0.8639 & 0.7425 & 0.6818 \\
\textbf{CPM} & 68.892 & 73.565 & 92.384 & 113.514 & 81.055 & 86.357 \\
\textbf{CPC} & 86552.665 & 87798.939 & 118514.791 & 131383.715 & 109158.834 & 126657.153 \\
\bottomrule
\bottomrule
\end{tabular}}%
\label{tab:tab3}
\vspace{-0.10in}
\end{table}
From Table 3, we first observe that click behaviors are very sparse in both training and testing sets. For example, the CTR of the 1458 training set is 0.7959 $\times 10^{-3}$, and the CTR of the testing set is 0.838$\times 10^{-3}$. It seriously weakens the performance of the CTR prediction model and reduces the credibility of the estimated value of each ad impression. Secondly, we observe that the average market prices (CPMs) of the training set and the testing set change greatly on three datasets, indicating that the RTB environment does have significant differences in different ad delivery periods. It may affect the performance of the static bidding strategy in the new ad delivery period. Furthermore, Figure 4 gives the statistics of various indicators in ten days of the 1458 dataset. We can see that there are large fluctuations in all kinds of statistics during ten delivery periods. Thus the static optimal bidding strategy derived from the historical training periods is likely unsuitable for the new ad delivery period.\par
\begin{figure}[t]
\centering
\includegraphics[width=0.32\textwidth]{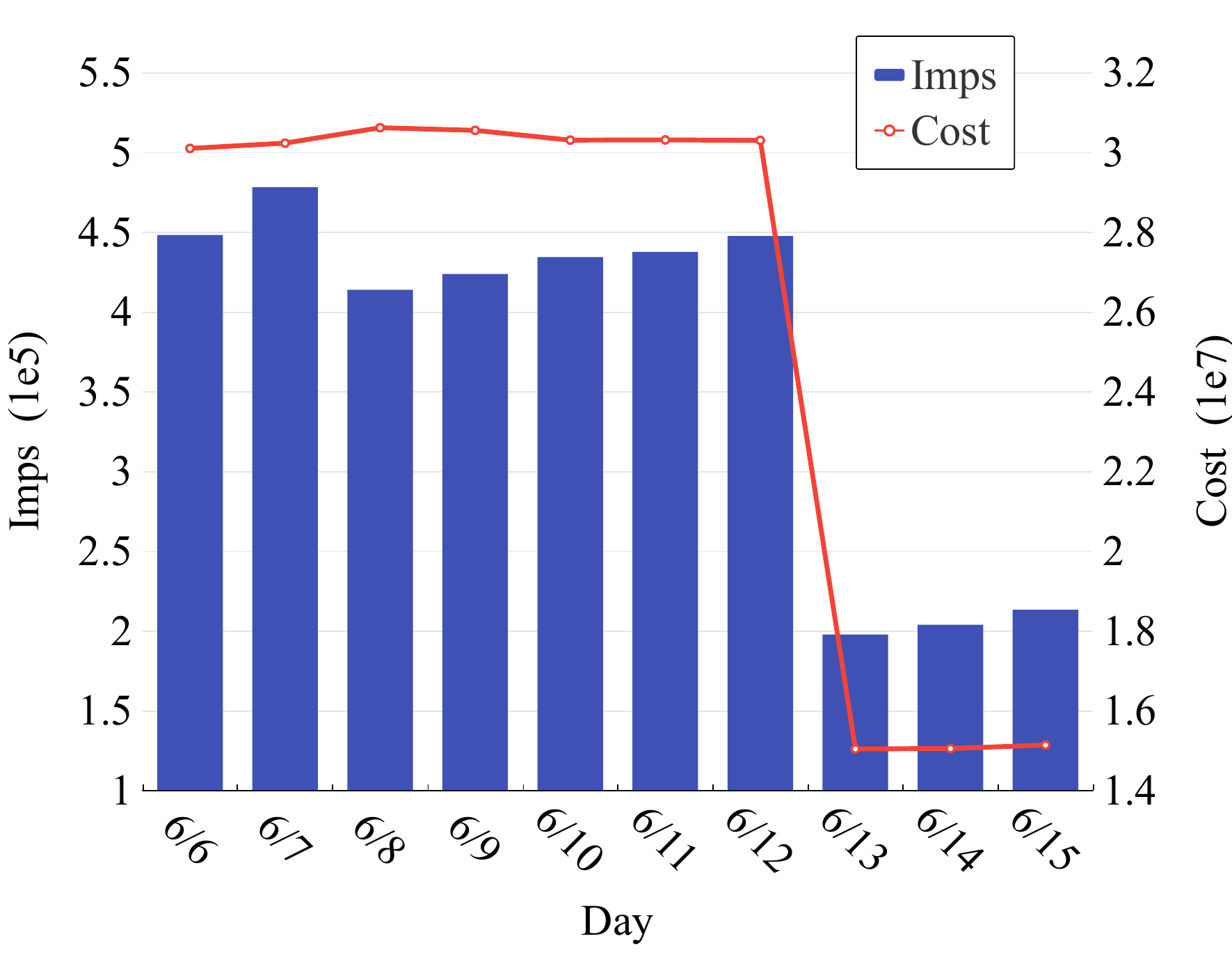}
\includegraphics[width=0.32\textwidth]{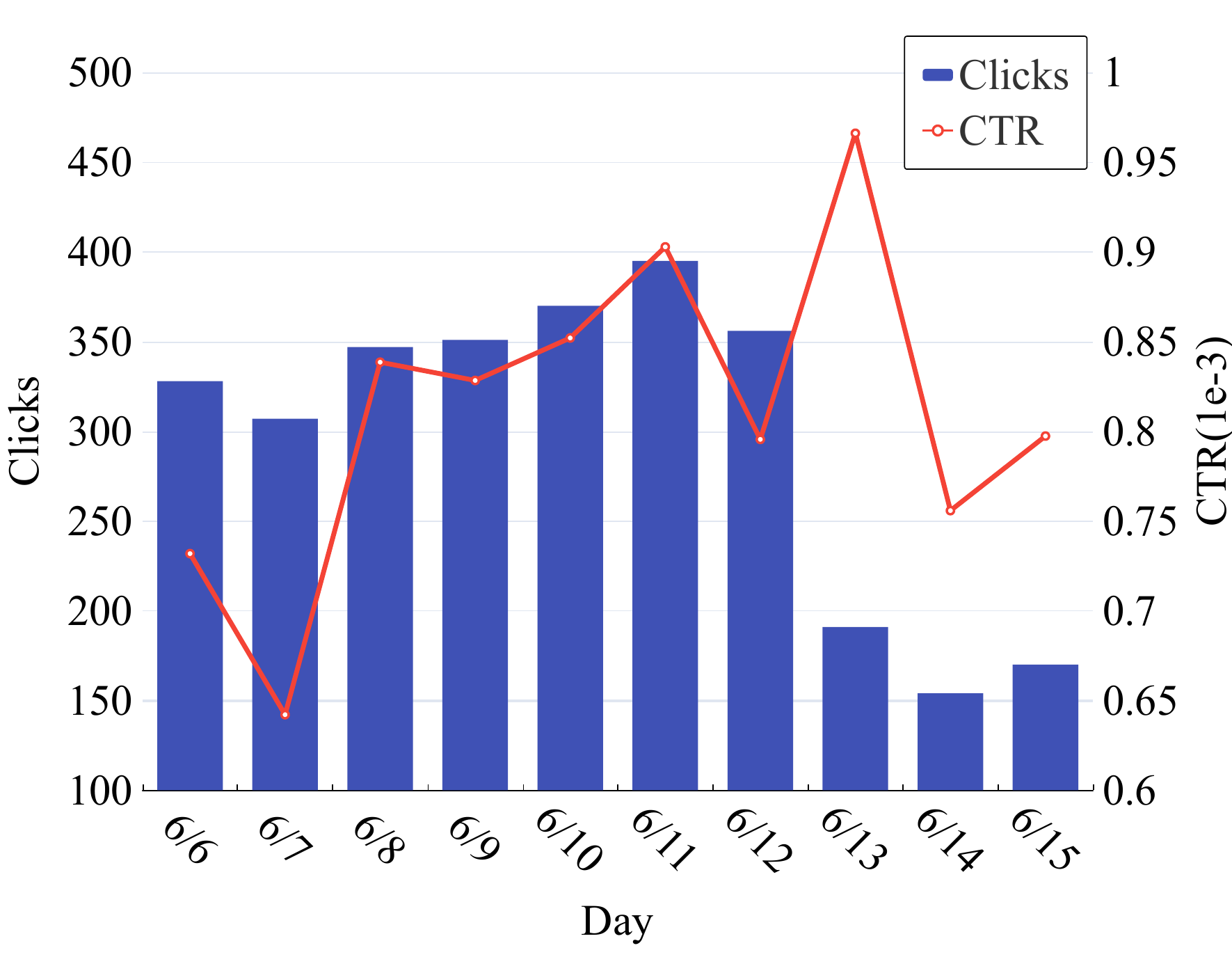}
\includegraphics[width=0.32\textwidth]{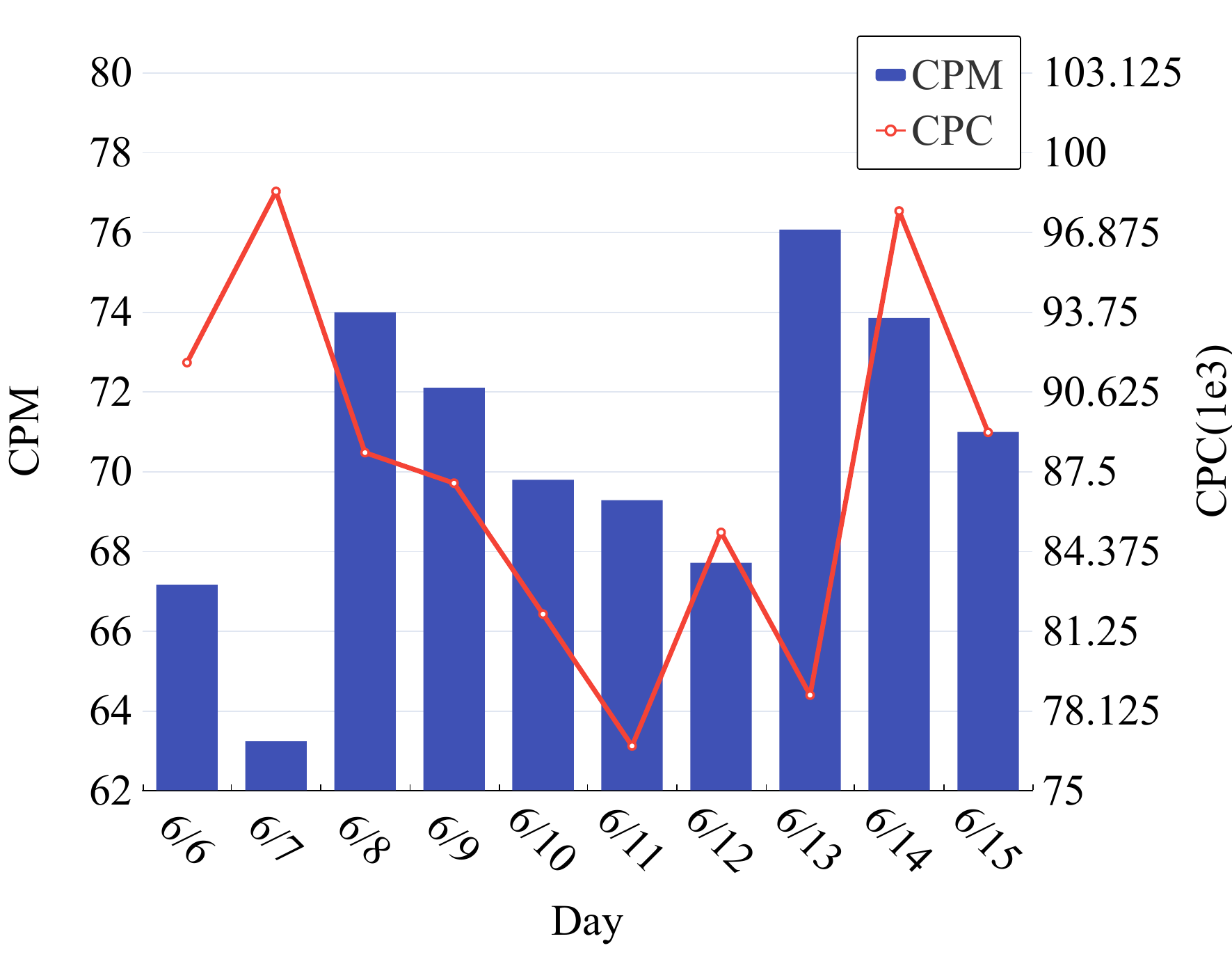}
\captionsetup{font={footnotesize}}
\caption{Statistics of Various Indicators in Ten Days of 1458 Dataset}
\label{fig:fig4}
\end{figure}
\begin{table}[t]
\centering
\caption{AUC Values of CTR Estimators}
\resizebox{0.4\textwidth}{!}{%
\begin{tabular}{ccc}
\toprule
\toprule
\textbf{Advertiser ID} & \textbf{Model} & \textbf{AUC}\\
\midrule
1458 & FNN & 0.7944\\
3358 & FM & 0.8675\\
3427 & FM & 0.8334\\
\bottomrule
\bottomrule
\end{tabular}}%
\label{tab:tab4}
\vspace{-0.10in}
\end{table}
\subsection{Click-Through Rate Prediction Model}
In this paper, all bidding strategies are based on the estimated pCTR of the ad impression to the advertiser. We use three representative models (LR \cite{ref6}, FM \cite{ref24}, and FNN \cite{ref25}) to train three CTR estimators for each advertiser and choose the best one as the estimator used in our experiments. Table 4 lists the CTR prediction models we choose for three advertisers and their AUC values. In our experiments, each advertiser’s bidding agent first computes the pCTR of the new arriving impression using its CTR estimator and decides the bidding price based on the pCTR. It should be noted that some click behaviors are quite inconsistent with the impressions’ pCTRs due to the uncertainty of user click behavior.\par
\subsection{Baseline Bidding Strategies}
In this subsection, we introduce some representative bidding strategies as baselines.\par
\begin{itemize}
\item \textbf{LIN}: The linear bidding strategy is defined as (3).\par
\item \textbf{RLB}: The bidding strategy is learned based on a model-based RL framework proposed in \cite{ref8}, which can directly select an optimal bidding price for each impression.\par
\item \textbf{DRLB}: The bidding strategy is learned based on a model-free RL framework proposed in \cite{ref11}. Primarily, it uses the DQN algorithm to train the optimal action selection policy that can help choose the regulating value to adjust each time slot’s bidding factor.\par
\item \textbf{OURS}: The bidding strategy is learned based on a model-free maximum entropy (SAC) RL framework proposed in this paper.\par
\end{itemize}
\subsection{Hyper-Parameters Setting}
In our algorithm, both Actor and four Critic networks use the feed-forward, fully connected neural network. Each network contains two hidden layers, and each hidden layer contains 128 neurons. The algorithm uses the Adam optimizer \cite{ref26} to optimize the neural network parameters, and the output layer uses the tanh activation function to constrain the output adjustment factor. We have described the detailed learning process of the algorithm in Section 4. Some hyper-parameters setting in our experiments are given in Table 5. It should be noted that the hyper-parameters setting of SAC mainly refers to the parameters setting in \cite{ref23}.\par
\begin{table}[t]
\centering
\caption{Key Hyper-parameter Implementation}
\resizebox{\textwidth}{!}{%
\begin{tabular}{ccl}
\toprule
\toprule
\textbf{Parameter} & \textbf{Setting} & \textbf{Description}\\
\midrule
$\gamma$ & 1 & Discount factor for TD in formula (15).\\
$\tau$ & 0.0005 & Soft-update parameter in formula (17).\\
$M$ & 1000000 & Size of replay buffer in Algorithm 1.\\
$N$ & 256 & Size of mini-batch in Algorithm 1.\\
$k$ & 30000 & Training once after $k$ pieces of experience has been stored in Algorithm 1.\\
$L$ & 128 & Number of rounds per training in Algorithm 1.\\
$d$ & 4 & Update frequency for soft updates in Algorithm 1.\\
\bottomrule
\bottomrule
\end{tabular}}%
\label{tab:tab5}
\vspace{-0.10in}
\end{table}
\section{Experimental Results}
\label{Experimental Results}
In this section, we first evaluate the performance of LIN (a typical static bidding strategy) on three datasets and discuss the impact of the RTB environment dynamics on the performance of LIN. Then, we compare our scheme with LIN and two dynamic RL-based bidding strategies (i.e., RLB and DRLB). Finally, we discuss the advantages and disadvantages of our scheme and LIN from the number of purchased impressions, the average market price, cost ratio, and the reasons for losing clicks. In order to evaluate the adaptability of the bidding strategy to budget changes, we set four different daily budgets for each set of experiments. Since the number of ad impressions and the actual cost vary from day to day in the dataset, we set the daily budget as 1/2, 1/4, 1/8, and 1/16 of the actual cost of each advertiser. Table 6 lists the actual costs of three advertisers on ten days.\par
\begin{table}[b]
\centering
\renewcommand\arraystretch{1.2}
\caption{Actual Daily Costs of three Advertisers on Ten Days($10^{-3}$ Chinese FEN)}
\resizebox{\textwidth}{!}{%
\begin{tabular}{ccccccccccc}
\toprule
\toprule
\multirow{2}[4]{*}{\textbf{Advertiser ID}}& \multicolumn{7}{|c|}{\textbf{Training set}} & \multicolumn{3}{c}{\textbf{Testing set}}\\
\cmidrule{2-11}
& \multicolumn{1}{|c}{\textbf{6/6}} & \textbf{6/7} & \textbf{6/8} & \textbf{6/9} & \textbf{6/10} & \textbf{6/11} & \multicolumn{1}{c|}{\textbf{6/12}} & \textbf{6/13} & \textbf{6/14} & \textbf{6/15}\\
\midrule
\textbf{1458} & \multicolumn{1}{|c}{30096630} & 30228554 & 30615541 & 30548604 & 30303929 & 30309883 & \multicolumn{1}{c|}{30297100} & 15036900 & 15045650 & 15133904\\
\textbf{3358} & \multicolumn{1}{|c}{17068590} & 17155542 & 16219705 & 14571538 & 40071957 & 23340047 & \multicolumn{1}{c|}{32515710} & 10864298 & 12143044 & 11152426\\
\textbf{3427} & \multicolumn{1}{|c}{30644030} & 23930230 & 29840853 & 32019674 & 31232220 & 30918866 & \multicolumn{1}{c|}{31654042} & 15185670 & 15325090 & 15845760\\
\bottomrule
\bottomrule
\end{tabular}}%
\label{tab:tab6}
\vspace{-0.10in}
\end{table}
\subsection{Performance of the Static Linear Bidding Strategy}
In this subsection, we discuss the impact of the dynamic RTB environment on the performance of the static bidding strategy. To this end, we perform LIN, the most representative static bidding strategy, on three datasets under four daily budget constraints. We run two groups of experiments for comparison. In the first group of experiments, we learn the optimal base bid to win the most clicks on the training set via a greedy heuristic algorithm. Then, we use the learned optimal base bid to calculate the bidding price for each ad impression on the testing set. As listed in Table 7, we record the results of the first group of experiments as \textbf{Base bid learned on Training Set}, which represents that the optimal base bid is learned on the training set and the clicks are obtained on the testing set.\par
In the second group of experiments, we assume that the testing set is known and learn the optimal base bid on the testing set. Then, we still use the learned optimal base bid to compute the bidding price for each impression on the testing set and record the results as \textbf{Base bid learned on Testing Set} since both optimal base bid and clicks are obtained on the testing set. The clicks received in the second group of experiments are the best results obtained by LIN under the environment of the testing periods. Therefore, we can analyze the adaptability of static bidding strategy to the environment changes quantitatively by comparing the click numbers obtained by the two groups of experiments.\par
\begin{table}[t]
\centering
\caption{Total Number of Clicks that the Linear Bidding Strategy Received on the Testing Set}
\resizebox{\textwidth}{!}{%
\begin{tabular}{cccccccccccccccccc}
    \toprule
    \toprule
    \multirowcell{2}[-0.5ex]{\textbf{Advertiser} \\ \textbf{ID}} & \multirowcell{2}[-0.5ex]{\textbf{Base bid} \\ \textbf{learned on}} & \multicolumn{4}{|c}{\textbf{1/2}} & \multicolumn{4}{c}{\textbf{1/4}} & \multicolumn{4}{c}{\textbf{1/8}} & \multicolumn{4}{c}{\textbf{1/16}} \\
    \cmidrule{3-18}
     & & \multicolumn{1}{|c}{\textbf{6/13}} & \textbf{6/14} & \textbf{6/15} & \multicolumn{1}{c|}{\textbf{Total}} & \textbf{6/13} & \textbf{6/14} & \textbf{6/15} & \multicolumn{1}{c|}{\textbf{Total}} & \textbf{6/13} & \textbf{6/14} & \textbf{6/15} & \multicolumn{1}{c|}{\textbf{Total}} & \textbf{6/13} & \textbf{6/14} & \textbf{6/15} & \textbf{Total} \\
    \midrule
    \multirow{2}[4]{*}{\textbf{1458}} & \multicolumn{1}{c|}{\textbf{Training set}} & 155 & 119 & 136 & \multicolumn{1}{c|}{410} & 96 & 81 & 89 & \multicolumn{1}{c|}{266} & 66 & 55 & 58 & \multicolumn{1}{c|}{179} & 39 & 40 & 47 & 126 \\
\cmidrule{2-2} & \multicolumn{1}{c|}{\textbf{Testing Set}} & 167 & 140 & 160 & \multicolumn{1}{c|}{467} & 103 & 107 & 117 & \multicolumn{1}{c|}{327} & 68 & 67 & 73 & \multicolumn{1}{c|}{208} & 39 & 44 & 49 & 132 \\
    \midrule
    \multirow{2}[4]{*}{\textbf{3358}} & \multicolumn{1}{c|}{\textbf{Training set}} & 68 & 79 & 82 & \multicolumn{1}{c|}{229} & 55 & 70 & 66 & \multicolumn{1}{c|}{191} & 48 & 62 & 60 & \multicolumn{1}{c|}{170} & 43 & 50 & 44 & 137 \\
\cmidrule{2-2} & \multicolumn{1}{c|}{\textbf{Testing Set}} & 67 & 78 & 82 & \multicolumn{1}{c|}{227} & 63 & 74 & 70 & \multicolumn{1}{c|}{207} & 54 & 63 & 61 & \multicolumn{1}{c|}{178} & 45 & 52 & 49 & 146 \\
    \midrule
    \multirow{2}[4]{*}{\textbf{3427}} & \multicolumn{1}{c|}{\textbf{Training set}} & 99 & 103 & 114 & \multicolumn{1}{c|}{316} & 80 & 82 & 90 & \multicolumn{1}{c|}{252} & 63 & 58 & 69 & \multicolumn{1}{c|}{190} & 49 & 47 & 54 & 150 \\
\cmidrule{2-2} & \multicolumn{1}{c|}{\textbf{Testing Set}} & 100 & 103 & 116 & \multicolumn{1}{c|}{319} & 84 & 84 & 92 & \multicolumn{1}{c|}{260} & 67 & 67 & 76 & \multicolumn{1}{c|}{210} & 57 & 52 & 61 & 170 \\
    \bottomrule
    \bottomrule
    \end{tabular}%
}%
\label{tab:tab7}
\vspace{-0.10in}
\end{table}
Obviously, all click numbers in the first group of experiments are lower than those in the second group of experiments, proving that the base bids learned on the historical training periods are not the best for the testing periods due to the dynamic changes of the RTB environment. Especially on the 1458 dataset, when we set the daily budget 1/2 and 1/4 of the cost, the click numbers are 410 and 266 in the first group of experiments, 12.20\% and 18.65\% lower than the ideal clicks (obtained in the second group of experiments). The results reveal that the RTB environment does change dramatically between the training and the testing periods and harms the performance of the static bidding strategy.\par
Moreover, we observe that the gaps between the actual and ideal clicks decrease with the daily budget declining. When we set the daily budget 1/16 of the cost, the gap is reduced to 6 clicks. Intuitively, we list the learned optimal base bids based on both training and testing sets in Table 8, demonstrating the optimal base bids fluctuating with the RTB environment changes. We find a common phenomenon that the optimal base bids decrease with the budget shrinking, which reveals a conservative bidding strategy is beneficial to maximize the overall clicks in the iPinYou dataset when the budget is insufficient. To sum up, through the experimental results in this subsection, we prove that the dynamic RTB environment does hinder the performance of the static bidding strategy and confirm the necessity of introducing a real-time adjustment scheme into the bidding strategy.\par
\begin{table}[t]
\centering
\caption{Optimal Base Bid(based on the maximum number of clicks)}
\resizebox{\textwidth}{!}{%
\begin{tabular}{ccccccccc}
\toprule
\toprule
\multirow{3}[5]{*}{\textbf{Advertiser ID}} & \multicolumn{4}{|c}{\textbf{1/2}} & \multicolumn{4}{|c}{\textbf{1/4}} \\
\cmidrule{2-9}
& \multicolumn{1}{|c|}{\textbf{Training set}} & \multicolumn{3}{c}{\textbf{Testing set}} & \multicolumn{1}{|c|}{\textbf{Training set}} & \multicolumn{3}{c}{\textbf{Testing set}} \\
\cmidrule{2-9}
& \multicolumn{1}{|c|}{\textbf{6/6-6/12}} & \textbf{6/13} & \textbf{6/14} & \textbf{6/15} & \multicolumn{1}{|c|}{\textbf{6/6-6/12}} & \textbf{6/13} & \textbf{6/14} & \textbf{6/15} \\
\midrule
\textbf{1458} & \multicolumn{1}{|c|}{298} & 184 & 173 & 194 & \multicolumn{1}{|c|}{111} & 64 & 59 & 62 \\
\textbf{3358} & \multicolumn{1}{|c|}{182} & 227 & 293 & 280 & \multicolumn{1}{|c|}{70} & 183 & 143 & 139 \\
\textbf{3427} & \multicolumn{1}{|c|}{212} & 228 & 207 & 220 & \multicolumn{1}{|c|}{80} & 92 & 82 & 94 \\
\midrule
\multirow{3}[5]{*}{\textbf{Advertiser ID}} & \multicolumn{4}{|c}{\textbf{1/8}} & \multicolumn{4}{|c}{\textbf{1/16}} \\
\cmidrule{2-9}
& \multicolumn{1}{|c|}{\textbf{Training set}} & \multicolumn{3}{c}{\textbf{Testing set}} & \multicolumn{1}{|c|}{\textbf{Training set}} & \multicolumn{3}{c}{\textbf{Testing set}} \\
\cmidrule{2-9}
& \multicolumn{1}{|c|}{\textbf{6/6-6/12}} & \textbf{6/13} & \textbf{6/14} & \textbf{6/15} & \multicolumn{1}{|c|}{\textbf{6/6-6/12}} & \textbf{6/13} & \textbf{6/14} & \textbf{6/15} \\
\midrule
\textbf{1458} & \multicolumn{1}{|c|}{48} & 35 & 33 & 33 & \multicolumn{1}{|c|}{25} & 23 & 25 & 21 \\
\textbf{3358} & \multicolumn{1}{|c|}{45} & 94 & 80 & 78 & \multicolumn{1}{|c|}{30} & 57 & 46 & 26 \\
\textbf{3427} & \multicolumn{1}{|c|}{44} & 50 & 49 & 54 & \multicolumn{1}{|c|}{27} & 33 & 32 & 36 \\
\bottomrule
\bottomrule
\end{tabular}}%
\label{tab:tab8}
\vspace{-0.10in}
\end{table}
\subsection{Performance Comparison of RL Bidding Strategies}
\begin{table}[!t]
\centering
\caption{Number of Clicks Received by Four Bidding Strategies under Different Budget Constraints}
\resizebox{\textwidth}{!}{%
\begin{tabular}{ccccccccc}
\toprule
\toprule
\multirow{2}[3]{*}{\textbf{Advertiser ID}} & \multicolumn{4}{|c}{\textbf{1/2}} & \multicolumn{4}{|c}{\textbf{1/4}}\\
\cmidrule{2-9}
& \multicolumn{1}{|c}{\textbf{LIN}} & \textbf{RLB} & \textbf{DRLB} & \textbf{OURS} & \multicolumn{1}{|c}{\textbf{LIN}} & \textbf{RLB} & \textbf{DRLB} & \textbf{OURS} \\
\midrule
\textbf{1458} & \multicolumn{1}{|c}{410} & 415 & 442 & \textbf{461} & \multicolumn{1}{|c}{266} & 294 & 294 & \textbf{296} \\
\textbf{3358} & \multicolumn{1}{|c}{229} & 221 & 228 & \textbf{235} & \multicolumn{1}{|c}{191} & 187 & 195 & \textbf{200} \\
\textbf{3427} & \multicolumn{1}{|c}{316} & 304 & 281 & \textbf{317} & \multicolumn{1}{|c}{252} & 246 & 234 & \textbf{263} \\
\midrule
\multirow{2}[3]{*}{\textbf{Advertiser ID}} & \multicolumn{4}{|c}{\textbf{1/8}} & \multicolumn{4}{|c}{\textbf{1/16}}\\
\cmidrule{2-9}
& \multicolumn{1}{|c}{\textbf{LIN}} & \textbf{RLB} & \textbf{DRLB} & \textbf{OURS} & \multicolumn{1}{|c}{\textbf{LIN}} & \textbf{RLB} & \textbf{DRLB} & \textbf{OURS} \\
\midrule
\textbf{1458} & \multicolumn{1}{|c}{179} & 176 & 172 & \textbf{184} & \multicolumn{1}{|c}{\textbf{126}} & 112 & 106 & 119 \\
\textbf{3358} & \multicolumn{1}{|c}{\textbf{170}} & 147 & 164 & 162 & \multicolumn{1}{|c}{\textbf{137}} & 112 & 132 & 131 \\
\textbf{3427} & \multicolumn{1}{|c}{190} & 198 & 164 & \textbf{202} & \multicolumn{1}{|c}{150} & 144 & 134 & \textbf{155} \\
\bottomrule
\bottomrule
\end{tabular}}%
\label{tab:tab9}
\vspace{-0.10in}
\end{table}
In this subsection, we evaluate our scheme with two typical RL-based bidding strategies. We still run each bidding strategy on three datasets under four budget constraints. Table 9 outlines the total number of clicks obtained by each bidding strategy in the testing periods. Also, we add the clicks received by LIN in Table 9 for comparison. First of all, we find that our scheme achieves the highest clicks in most experiments among three RL-based bidding strategies, with significant advantages compared with RLB and DRLB. Among three RL-based bidding strategies, RLB and DRLB get the least clicks respectively in the six experiments. Only when the budget is set to 1/8 of cost, the click number obtained by our scheme is slightly lower than that of DRLB on the 3358 dataset. In addition, our strategy is also superior to LIN. Especially when the daily budgets are 1/2 and 1/4 of the cost, our scheme can obtain 461 and 296 clicks on the 1458 dataset, 12.44\% and 11.28\% higher than LIN. Similar results are found on the 3358 and 3427 datasets. Such results show that our scheme can adjust the bidding price adaptively according to the RTB environment in real-time so that the bidding price of each ad impression can match its environment as much as possible. Furthermore, the results also prove that it is feasible to introduce the adjustment factor into the bidding function of LIN.\par
\begin{table}[!b]
\centering
\caption{Sum of pCTR Won by Four Bidding Strategies under Different Budget Constraints}
\resizebox{\textwidth}{!}{%
\begin{tabular}{ccccccccc}
\toprule
\toprule
\multirow{2}[3]{*}{\textbf{Advertiser ID}} & \multicolumn{4}{|c}{\textbf{1/2}} & \multicolumn{4}{|c}{\textbf{1/4}}\\
\cmidrule{2-9}
& \multicolumn{1}{|c}{\textbf{LIN}} & \textbf{RLB} & \textbf{DRLB} & \textbf{OURS} & \multicolumn{1}{|c}{\textbf{LIN}} & \textbf{RLB} & \textbf{DRLB} & \textbf{OURS} \\
\midrule
\textbf{1458} & \multicolumn{1}{|c}{330.07} & 355.02 & 350.84 & \textbf{368.22} & \multicolumn{1}{|c}{230.91} & 275.09 & 246.96 & \textbf{249.90} \\
\textbf{3358} & \multicolumn{1}{|c}{307.14} & 294.42 & 306.89 & \textbf{309.96} & \multicolumn{1}{|c}{258.49} & 246.83 & \textbf{265.28} & 259.36 \\
\textbf{3427} & \multicolumn{1}{|c}{\textbf{413.49}} & 393.03 & 353.03 & 404.86 & \multicolumn{1}{|c}{\textbf{328.89}} & 313.06 & 321.43 & 327.22 \\
\midrule
\multirow{2}[3]{*}{\textbf{Advertiser ID}} & \multicolumn{4}{|c}{\textbf{1/8}} & \multicolumn{4}{|c}{\textbf{1/16}}\\
\cmidrule{2-9}
& \multicolumn{1}{|c}{\textbf{LIN}} & \textbf{RLB} & \textbf{DRLB} & \textbf{OURS} & \multicolumn{1}{|c}{\textbf{LIN}} & \textbf{RLB} & \textbf{DRLB} & \textbf{OURS} \\
\midrule
\textbf{1458} & \multicolumn{1}{|c}{177.72} & 191.56 & 179.09 & \textbf{182.50} & \multicolumn{1}{|c}{\textbf{140.29}} & 129.71 & 103.98 & 115.80 \\
\textbf{3358} & \multicolumn{1}{|c}{\textbf{225.53}} & 188.5 & 217.58 & 213.97 & \multicolumn{1}{|c}{\textbf{181.88}} & 140.55 & 174.79 & 163.80 \\
\textbf{3427} & \multicolumn{1}{|c}{249.77} & 237.47 & 230.49 & \textbf{251.46} & \multicolumn{1}{|c}{184.63} & 182.13 & 167.94 & \textbf{188.92} \\
\bottomrule
\bottomrule
\end{tabular}}%
\label{tab:tab10}
\vspace{-0.10in}
\end{table}
Table 10 presents the sum of pCTR won by each bidding strategy in the testing periods. Firstly, we observe that, similar to the number of clicks, our scheme achieves the largest pCTR in most experiments among the three RL-based bidding strategies. Moreover, we note that when the budget is 1/4 of cost, the pCTR won by our scheme is marginally lower than that of DRLB on the 3358 dataset while our scheme obtained more clicks. The same happens on the 1458 dataset when the budget is set to 1/8 and 1/16. This results show that our scheme purchased more valuable ad impressions that generate clicks.\par
\subsection{Detailed Comparison of Our Scheme with LIN}
In the last subsection, we discuss the advantages and disadvantages of our scheme and LIN in detail. First, Table 11 gives the numbers of impressions purchased by two strategies on the testing set. From this table, we observe that our scheme buys more ad impressions than LIN in eight experiments because our scheme can adjust its bidding price for every auctioned impression dynamically according to the environment of the testing periods. In contrast, LIN is a static bidding strategy in which the optimal base bid is only learned from the training periods and cannot be adjusted according to the real-time RTB environment. If the learned optimal base bid is significantly lower than that in the testing periods, it will lead the advertiser to bid with low price and lose many available impressions. As a result, there will be a massive budget surplus. For example, all budgets have not been spent out in four experiments on the 3427 dataset.\par
\begin{table}[t]
\centering
\caption{Numbers of Ad Impressions Purchased by LIN and Our Scheme}
\resizebox{\textwidth}{!}{%
\begin{tabular}{ccccccccc}
\toprule
\toprule
\multirow{2}[3]{*}{\textbf{Advertiser ID}} & \multicolumn{2}{|c}{\textbf{1/2}} & \multicolumn{2}{|c}{\textbf{1/4}} & \multicolumn{2}{|c}{\textbf{1/8}} & \multicolumn{2}{|c}{\textbf{1/16}} \\
\cmidrule{2-9}
& \multicolumn{1}{|c}{\textbf{LIN}} & \textbf{OURS} & \multicolumn{1}{|c}{\textbf{LIN}} & \textbf{OURS} & \multicolumn{1}{|c}{\textbf{LIN}} & \textbf{OURS} & \multicolumn{1}{|c}{\textbf{LIN}} & \textbf{OURS} \\
\midrule
\textbf{1458} & \multicolumn{1}{|c}{\textbf{337559}} & 327855 & \multicolumn{1}{|c}{186698} & \textbf{187135} & \multicolumn{1}{|c}{110719} & \textbf{119257} & \multicolumn{1}{|c}{\textbf{65456}} & 60501\\
\textbf{3358} & \multicolumn{1}{|c}{158301} & \textbf{172912} & \multicolumn{1}{|c}{74835} & \textbf{79779} & \multicolumn{1}{|c}{\textbf{44594}} & 43978 & \multicolumn{1}{|c}{\textbf{22662}} & 19720\\
\textbf{3427} & \multicolumn{1}{|c}{309703} & \textbf{314510} & \multicolumn{1}{|c}{161316} & \textbf{171027} & \multicolumn{1}{|c}{84099} & \textbf{93411} & \multicolumn{1}{|c}{40969} & \textbf{46986}\\
\bottomrule
\bottomrule
\end{tabular}}%
\label{tab:tab11}
\vspace{0.10in}
\end{table}
Table 12 shows the average market prices of our strategy and LIN. In most cases, the results reflect that our scheme’s average market prices are higher than those of LIN since our scheme adopts an aggressive bidding strategy. Therefore, our scheme can work well when the budget is sufficient. However, it performs marginally poorly when the budget is seriously inadequate. In particular, when the budget is only 1/16 of the cost, the click numbers on the 1458 and 3358 datasets are slightly lower than those of LIN. In the following Table 14, we further analyze the main reason why advertisers lose clicks.\par
\begin{table}[!t]
\centering
\caption{Average Costs of Buying an Ad Impression in LIN and Our Scheme ($10^{-3}$ Chinese FEN)}
\resizebox{\textwidth}{!}{%
\begin{tabular}{ccccccccc}
\toprule
\toprule
\multirow{2}[3]{*}{\textbf{Advertiser ID}} & \multicolumn{2}{|c}{\textbf{1/2}} & \multicolumn{2}{|c}{\textbf{1/4}} & \multicolumn{2}{|c}{\textbf{1/8}} & \multicolumn{2}{|c}{\textbf{1/16}}\\
\cmidrule{2-9}
& \multicolumn{1}{|c}{\textbf{LIN}} & \textbf{OURS} & \multicolumn{1}{|c}{\textbf{LIN}} & \textbf{OURS} & \multicolumn{1}{|c}{\textbf{LIN}} & \textbf{OURS} & \multicolumn{1}{|c}{\textbf{LIN}} & \textbf{OURS}\\
\midrule
\textbf{1458} & \multicolumn{1}{|c}{66.975} & \textbf{68.806} & \multicolumn{1}{|c}{\textbf{60.547}} & 60.406 & \multicolumn{1}{|c}{\textbf{51.048}} & 47.393 & \multicolumn{1}{|c}{43.174} & \textbf{46.710}\\
\textbf{3358} & \multicolumn{1}{|c}{96.821} & \textbf{97.792} & \multicolumn{1}{|c}{91.488} & \textbf{96.040} & \multicolumn{1}{|c}{90.143} & \textbf{97.002} & \multicolumn{1}{|c}{92.571} & \textbf{108.264}\\
\textbf{3427} & \multicolumn{1}{|c}{72.888} & \textbf{73.535} & \multicolumn{1}{|c}{62.219} & \textbf{67.617} & \multicolumn{1}{|c}{54.940} & \textbf{62.032} & \multicolumn{1}{|c}{46.198} & \textbf{52.563}\\
\bottomrule
\bottomrule
\end{tabular}}%
\label{tab:tab12}
\vspace{-0.10in}
\end{table}
Furthermore, we discuss the cost ratio of our scheme and LIN, where the cost ratio is the ratio of the actual cost to the budget. Ideally, we hope to obtain the most clicks in an ad delivery period within a given budget. In this subsection, we only give the experimental results on the 3427 dataset in Table 13. First, we observe that the cost ratios of LIN are significantly lower than those of our scheme. Because LIN is a static bidding strategy, the bidding agent cannot adjust its base bid according to the environment in real-time. Specifically, on the 3427 dataset, the learned base bid is relatively low compared to the environment of testing periods, leading to the bidding price for an impression usually lower than its market prices. Therefore, the advertiser loses many impressions and has a lot of money left at the end of the ad delivery period. On the other hand, the advertiser 3427 loses substantial impressions that may bring clicks, which will hurt the advertiser’s revenue in the testing periods.\par
\begin{table}[!b]
\centering
\caption{Cost Ratios and Winning Impressions Obtained by LIN and Our Scheme on 3427}
\resizebox{\textwidth}{!}{%
\begin{tabular}{ccccccccc}
\toprule
\toprule
& \multicolumn{2}{|c}{\textbf{1/2}} & \multicolumn{2}{|c}{\textbf{1/4}} & \multicolumn{2}{|c}{\textbf{1/8}} & \multicolumn{2}{|c}{\textbf{1/16}}\\
\cmidrule{2-9}
& \multicolumn{1}{|c}{\textbf{LIN}} & \textbf{OURS} & \multicolumn{1}{|c}{\textbf{LIN}} & \textbf{OURS} & \multicolumn{1}{|c}{\textbf{LIN}} & \textbf{OURS} & \multicolumn{1}{|c}{\textbf{LIN}} & \textbf{OURS}\\
\midrule
\textbf{Cost ratio} & \multicolumn{1}{|c}{97.392\%} & \textbf{99.782\%} & \multicolumn{1}{|c}{86.607\%} & \textbf{99.787\%} & \multicolumn{1}{|c}{78.286\%} & \textbf{99.999\%} & \multicolumn{1}{|c}{65.326\%} & \textbf{85.244\%}\\
\textbf{Win imps} & \multicolumn{1}{|c}{309703} & \textbf{314510} & \multicolumn{1}{|c}{161316} & \textbf{171027} & \multicolumn{1}{|c}{84099} & \textbf{93411} & \multicolumn{1}{|c}{40969} & \textbf{46986}\\
\bottomrule
\bottomrule
\end{tabular}}%
\label{tab:tab13}
\vspace{0.10in}
\end{table}
In contrast, our scheme can get higher cost ratios under various cases due to supporting adjust the bidding price at the impression-grained level. In addition, we note that the cost ratio of LIN decreases as the budget shrinks. This happens because the optimal base price learned by LIN is reduced with the budget narrows to capture the impressions of the whole period as much as possible, avoiding the budget being spent out in advance. According to the bidding function of LIN, the lower the base bid is, the lower the bidding price is. Therefore, the number of impressions successfully purchased by LIN is greatly reduced. When the budget is 1/16 of the cost, only 65.33\% of the budget has been spent by LIN.\par
In RTB, there are two reasons why advertisers lose clicks. One is that the bidding price is less than the market price, and the other is that the advertiser has no money to buy impressions due to its budget has been wiped out in advance, resulting in losing all subsequent impressions. In RTB, we call the latter case as \textbf{early stop}. Here, we take the experimental results on the 1458 dataset to make detailed statistics of why LIN and our strategy lose the clicks. Table 14 presents the missing click numbers caused by each reason. We observe that when the budgets are 1/2, 1/4, and 1/8 of the cost, LIN loses more clicks than our scheme for early stop, but it loses fewer clicks than ours for its bidding price lower than the market price.\par
\begin{table}[t]
\centering
\caption{Number of Lost Clicks and Reason in 1458 Dataset}
\resizebox{\textwidth}{!}{%
\begin{tabular}{ccccccccc}
\toprule
\toprule
\multirow{2}[3]{*}{\textbf{Reason}} & \multicolumn{2}{|c}{\textbf{1/2}} & \multicolumn{2}{|c}{\textbf{1/4}} & \multicolumn{2}{|c}{\textbf{1/8}} & \multicolumn{2}{|c}{\textbf{1/16}}\\
\cmidrule{2-9}
& \multicolumn{1}{|c}{\textbf{LIN}} & \textbf{OURS} & \multicolumn{1}{|c}{\textbf{LIN}} & \textbf{OURS} & \multicolumn{1}{|c}{\textbf{LIN}} & \textbf{OURS} & \multicolumn{1}{|c}{\textbf{LIN}} & \textbf{OURS}\\
\midrule
\textbf{Early stop} & \multicolumn{1}{|c}{89} & \textbf{11} & \multicolumn{1}{|c}{175} & \textbf{119} & \multicolumn{1}{|c}{158} & \textbf{121} & \multicolumn{1}{|c}{\textbf{67}} & 105\\
\midrule
\makecell*[c]{\textbf{Bid lower than}\\\textbf{market price}}& \multicolumn{1}{|c}{\textbf{16}} & 43 & \multicolumn{1}{|c}{\textbf{74}} & 100 & \multicolumn{1}{|c}{\textbf{178}} & 210 & \multicolumn{1}{|c}{322} & \textbf{291}\\
\bottomrule
\bottomrule
\end{tabular}}%
\label{tab:tab14}
\vspace{-0.10in}
\end{table}
The reason for the above results is that the optimal base bid learned by LIN in 1458 training periods is obviously high for the environment of the testing periods. Consequently, LIN’s bidding price for each impression during the testing periods is much higher than its market price, resulting in LIN winning a large number of ad impressions. Among the ad impressions that LIN has bought, more of them are no-click impressions. Buying these useless ad impressions wastes a lot of money and can easily cause advertisers to spend out their budgets in advance and cannot purchase subsequent ad impressions. Our scheme introduces a dynamic adjustment factor into LIN. The bidding price of each impression can be adjusted dynamically to match the real-time environment of the testing periods, avoiding spending the budget out too quickly or too slowly. When the budget is set to 1/16 of the cost, in LIN, the optimal base bid learned on the training set fits the environment of testing periods. Hence, the number of lost clicks caused by budget depletion reduces significantly.\par
Based on the above analyses, we can draw the following conclusions. Firstly, the environment dynamics do have a negative impact on the performance of the static bidding strategy. Still, it has little impact on our dynamic bidding strategy based on maximum entropy RL. Secondly, the new bidding function designed by us is practical, which can adjust the bidding price for each impression according to the real-time environment by introducing an adjustment factor into LIN. Finally, learning the optimal adjustment factor generation policy utilizing the maximum entropy RL is superior to other RL models.\par
\section{Conclusions}
\label{Conclusions}
In this paper, we focus on using model-free RL to optimize advertisers’ bidding strategies. Specifically, we first design a new bidding function that uses LIN to compute the base price of each ad impression, and adjusts the base price to fit the real-time RTB environment by introducing a bidding adjustment factor. To this end, we model the adjustment factor decision as an MDP, and then use the stochastic policy SAC to solve the optimal adjustment factor generation policy. Unlike the widely used deterministic model-free RL algorithm, SAC can address the problem of multiple optimal actions for a single impression brought by GSP mechanism. Secondly, SAC also extends the scope of the RL agent to explore the optimal action, enabling the algorithm to converge to the global optimum more quickly. In particular, we design a new reward function that enables the RL agent to better learn the optimal action by comparing it with the results of bidding using LIN, thus maximizing the probability of the optimal action being selected. Finally, we validate our improvements on lots of experiments. The work in this paper focuses on bid optimization for a single advertiser. In real RTB applications, advertising platforms usually need to optimize the total revenue of multiple advertisers, and we hope to use reinforcement learning solutions to solve this challenge in the future. Undoubtedly, this will be a more valuable work.\par
\section*{Acknowledgements}
This work was supported in part by 1) the National Natural Science Foundation of China under Grant 61202445; 2) the Fundamental Research Funds for the Central Universities under Grant ZYGX2016J096.\par

\end{document}